\documentclass[journal]{IEEEtran}
\usepackage{textcomp}

\usepackage{algorithmic}
\usepackage{algorithm}
\usepackage{array}
\usepackage[caption=false,font=normalsize,labelfont=sf,textfont=sf]{subfig}
\usepackage{textcomp}
\usepackage{stfloats}
\usepackage{url}
\usepackage{verbatim}
\usepackage{graphicx}
\usepackage{amsmath}
\usepackage{cite}

\usepackage{booktabs}

\begin{document}

\title{Bridging Data Trials and Task Barriers: A Unified Framework for Sketch Biometric Identification}

\author{
    Decheng Liu,
    Bin Hu,
    Xinbo Gao,~\IEEEmembership{Fellow,~IEEE,}
    Dawei Zhou,
    Chunlei Peng,~\IEEEmembership{Member,~IEEE,}
    Nannan Wang,~\IEEEmembership{Senior~Member,~IEEE,}
    and Ruimin Hu 

}



\maketitle

\begin{abstract}
Different from existing cross-modality identification tasks (e.g., heterogeneous face recognition, sketch re-identification, etc.), we introduce a novel yet practical setting for these related identification tasks, named \textbf{sketch biometric identification}, which aims to continually train a unified model across different data domains, even diverse identification tasks. Sketch biometric identification faces challenges, including scarce real sketch data, high annotation costs, privacy risks, and insufficient generalization ability of cross-task models. Existing methods usually rely on limited real data or single-task optimization, making it difficult to effectively address the joint challenges of cross-modality and cross-task. This paper proposes a unified framework that integrates efficient synthetic sketch generation and task-sequential continual learning. First, we design an efficient pipeline to generate a large-scale and high-quality synthetic person and face sketch data, which significantly reduces costs and avoids privacy risks. Meanwhile, we enhance the model's robustness by fusing real data. Second, we construct a universal unified framework for sketch biometric identification, which adopts a task-sequential training strategy: the model first completes sketch person re-identification learning on the person dataset; subsequently, it maintains the acquired person recognition capability through a trusted sample replay technique and seamlessly performs incremental training on the face dataset. This enables a single model to simultaneously handle the cross-task capabilities of multiple sketch biometric identification tasks. To support the study of the mentioned sketch biometric identification, we built a new large-scale benchmark, SketchUnified-BioID, with several practical evaluation protocols. Sufficient experimental results prove the proposed unified framework achieves state-of-the-art performance on the two representative tasks (e.g., heterogeneous face recognition, sketch re-identification, etc.). We also find that the unified framework can effectively capture more intrinsic, task-shared discriminative features across multiple tasks, thus enabling robust generalization across diverse sketch biometrics scenarios. The source code and models are publicly available at https://github.com/sHanbIgsUn/UFSB.
\end{abstract}


\section{Introduction}
\begin{figure}[t]
  \centering
  
   \includegraphics[width=0.6\linewidth]{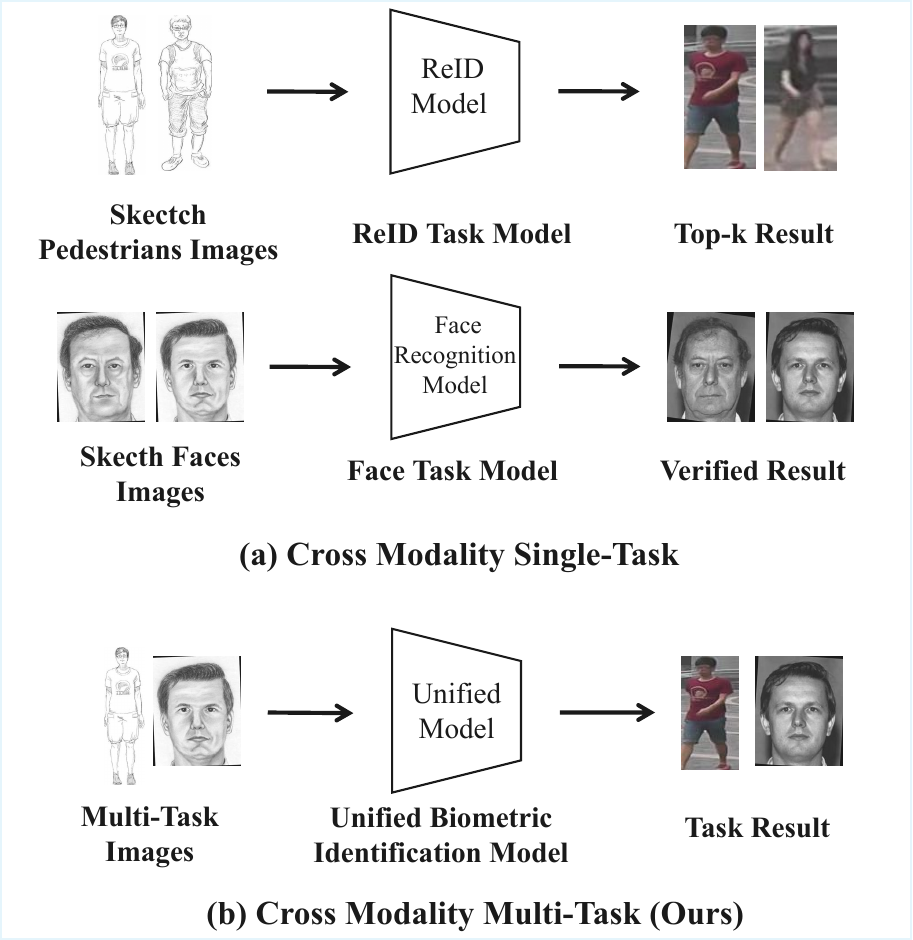}
    
   \caption{The top section shows the “single-task inference pipeline”, independent models for separate sketch biometric tasks. The bottom section presents our \textbf{Unified Sketch Biometric Identification model}, which uses a single model to handle multiple sketch biometric tasks via sequential learning, eliminating redundant deployment and mitigating catastrophic forgetting.}
   \label{fig:onecol}
\end{figure}
\IEEEPARstart{P}{erson} re-identification (ReID) has achieved remarkable progress in closed-set scenarios, where models are trained on static datasets with fixed modalities and tasks. Among its subfields, cross-modal sketch-photo ReID has garnered significant attention for its practical value in security and protection monitoring and forensic evidence \cite{1,4,22dkp}. Existing works in this area, however, remain confined to single-task paradigms: sketch-pedestrian ReID methods \cite{9,11} focus solely on aligning full-body sketch and photo features, while sketch-face ReID approaches \cite{1,6} only optimize for facial biometric matching. These methods excel in their respective tasks but fail to address the need for unified biometric identification across multiple sketch-based tasks.  

Meanwhile, lifelong person re-identification (LReID) has emerged to handle non-stationary data streams, aiming to mitigate catastrophic forgetting while acquiring new knowledge \cite{24pseudo,29}. Despite its success in visible-modal tasks, LReID research has not extended to sketch modalities. Existing LReID methods lack cross-modal optimization for sketch-photo pairs, leading to poor adaptability when faced with the unique challenges of sketch data. As real-world scenarios demand models that can simultaneously handle pedestrian and facial sketch identification, the gap between single-task sketch ReID and multi-task lifelong learning becomes increasingly prominent.  

To bridge this critical gap between fragmented single-task models and the holistic demands of real-world applications, we formally define and introduce a novel yet highly practical task: \textbf{Unified Cross-Modal Sketch Biometric Identification.} Unlike conventional approaches that deploy specialized models for isolated tasks (e.g., one for face sketches, another for full-body pedestrian sketches), our unified task requires a single, versatile model capable of processing any input sketch, regardless of whether it depicts a face, a full-body person, or potentially other biometric traits like a palmprint. The model's objective is then to perform accurate identification against a comprehensive gallery containing corresponding photographic records of these diverse biometrics.

This paradigm shift is motivated by realistic forensic and security scenarios. For instance, consider an investigation where an eyewitness provides multiple, partial descriptions of a suspect: a detailed facial sketch drawn at a police station, and a rough full-body sketch captured from a distant CCTV angle. A traditional system would require running these sketches through two separate, specialized pipelines. In contrast, our unified framework allows both sketches to be processed by the same model, enabling a consolidated search across a unified database of suspect photos. This not only streamlines deployment and reduces computational overhead but also holds the potential to leverage shared, intrinsic features across different biometric modalities for more robust identification. \emph{To the best of our knowledge, this work presents the first exploration of such a unified sketch biometric identification task,} which simultaneously handles sketch-pedestrian and sketch-face ReID—and is extensible to other sketch-based biometrics—while learning incrementally from non-stationary data streams.

Despite the practical value of this unified task, two critical challenges hinder its realization: \textbf{Scarcity of sketch data:} Real sketch datasets suffer from limited scale and high annotation costs, restricting model training and generalization. \textbf{Catastrophic forgetting in cross-task lifelong learning:} When learning a new biometric task, models tend to overwrite discriminative knowledge of the old task, leading to drastic performance degradation on previously learned tasks. This issue is exacerbated by the distinct feature distributions of pedestrian and face sketches. Existing methods fail to address these dual challenges, leaving a critical gap in unified sketch biometric identification. Single-task sketch ReID methods only optimize for a single biometric task and do not consider cross-task transfer. LReID approaches focus on visible-modal data and lack cross-modal optimization for sketch-photo pairs, making their performance far from practical requirements when adapted to sketch tasks.

To fill this gap, we propose a unified framework that integrates efficient synthetic sketch generation and task-sequential continual learning. Specifically, we first adopt a fast synthesis pipeline to construct two large-scale synthetic sketch datasets, addressing the scarcity of real sketch data. For cross-task lifelong learning, our method has two main modules. The Sketch Biometric Identification Foundation Module (SIM) uses a ClipReID backbone and a Joint Maximum Mean Discrepancy (JMMD) loss to align sketch and photo features. The Multi-Task Preservation Module (MPM) uses conformal prediction to pick low-uncertainty exemplars for replay; this reduces forgetting while avoiding large exemplar stores. Tasks are trained sequentially, enabling the model to accumulate multi-task knowledge incrementally.  

The main contributions of our paper are summarized as follows:
\begin{enumerate}
    \item A novel unified task of cross-modal sketch biometric identification: We are the first to define and address the task of simultaneously handling sketch-pedestrian and sketch-face ReID (and extendable to other sketch-based biometric tasks), filling the gap between single-task sketch ReID and multi-task lifelong learning.  
  \item A large-scale unified benchmark SketchUnified-BioID: This benchmark integrates real datasets and synthetic datasets, with fully non-overlapping identities across tasks, large-scale samples, and a dataset composed of real and synthetic images. The synthetic portion supplements scarce real sketches while test identities remain strictly disjoint to preserve objective evaluation.  
  \item Efficient multi-task preservation and cross-modal alignment: The proposed MPM module enables zero-shot, efficient sample selection to retain old-task knowledge.
  \item Extensive experimental results demonstrate that our framework achieves state-of-the-art performance. We find the interesting phenomenon that the unified framework can effectively capture more intrinsic, task-shared discriminative features across multiple tasks, thus enabling robust generalization across diverse sketch biometrics scenarios.
\end{enumerate}

\section{Related Work}
\subsection{Face sketch recognition.}

Face sketch recognition methods can be categorized into several technical paradigms:
\emph{Handcrafted synthesis based methods:} early works laid foundations via MRF \cite{1}, sampling optimization \cite{4}, and multi-style datasets \cite{5}, establishing basic sketch-photo mapping rules through manual feature engineering.
\emph{Deep learning based methods:} recent advances leverage CNN architectures for semantic alignment. Mittal et al. \cite{2,3} integrated transfer learning with attribute-guided saliency to enhance cross-modal feature transferability, while Iranmanesh et al. \cite{6} proposed a coupled deep network that embeds facial attributes to mitigate texture loss between sketch and photo. Lightweight CNN variants \cite{7} further improved computational efficiency.
\emph{Graph representation based methods:} Peng P. et al. \cite{8} strengthened cross-modal correlations via sparse graphical representation, capturing structural relationships between sketch and photo features.
\emph{Generative modeling based methods:} Fu et al. \cite{DVG} formulated heterogeneous face recognition as a dual variational generation problem, leveraging contrastive learning to produce domain-invariant embeddings across multiple modalities including sketch-photo. In parallel, Fu et al. \cite{SAFE} introduced a lightweight pixel-wise hallucination framework with 3D-assisted shape alignment, achieving efficient spectrum translation with minimal computational overhead.
Despite progress across these paradigms, all methods remain limited to single-task scenarios or require modality-specific architectures; our work extends this to unified cross-task sketch biometrics with sequential learning capability.

\subsection{Sketch person re-identification.}
Sketch person ReID focuses on domain alignment and data efficiency: early methods explored adversarial adaptation \cite{9} and embedding optimization \cite{18,19} to narrow the sketch-photo gap. Recent advances leverage Transformer and pre-trained models for better generalization: Chen et al. \cite{11} designed an asymmetrical disentanglement module in Transformers to decouple style and content features, while Chen et al. \cite{12} proposed a modality-agnostic architecture that unifies sketch and photo feature spaces. Lin et al. \cite{10BDG} constructed a multi-style sketch dataset and integrated a fusion module to handle subjective drawing variations. Zhang et al. \cite{15,16,17} further addressed data scarcity via embedding expansion, middle modality learning, and dual-semantic consistency. However, these works lack cross-task transfer capability; our framework fills this gap by unifying face and pedestrian sketch tasks.

\subsection{Lifelong Person Re-Identification.}
\emph{Knowledge distillation and consolidation:} early approaches leveraged knowledge distillation to transfer old model knowledge without exemplar storage. LSTKC++ \cite{LSTKC} developed long short-term knowledge decomposition with old knowledge rectification, while PAEMA \cite{PAEMA} introduced prompt-guided adaptive exponential moving average for dynamic consolidation. Pseudo task frameworks \cite{24pseudo} further addressed task-wise domain gaps by mapping new features into old task spaces.
\emph{Prototype and distribution modeling:} recent breakthroughs focus on capturing statistical characteristics of past tasks. DKP \cite{22dkp} proposed instance-level distribution modeling to transform diversity into identity-level prototypes for fine-grained transfer. DASK \cite{23dask} advanced this by designing adaptive style kernels to rehearse historical distributions, eliminating exemplar storage while maintaining anti-forgetting capacity.
\emph{Compatible representation and generalization:} to address practical deployment challenges, Bi-C2R \cite{BIC2R} introduced bidirectional continual compatible representation for re-indexing free inference. AKA\cite{27aka} and MEGE \cite{28MEGE} endowed models with knowledge operation capabilities via graph networks, jointly preventing forgetting while improving generalization to unseen domains. CLUDA-ReID \cite{26CLUDA} coordinated anti-forgetting with adaptation for streaming unlabeled data scenarios.
Despite progress, all methods remain limited to single-modal pedestrian ReID without cross-modal optimization for sketch-photo pairs. Our work extends LReID to unified cross-task sketch biometrics, integrating cross-modal alignment into the distribution-aware framework.

\subsection{Replay-based Continual Learning}
Replay-based continual learning has two classic paradigms: early exemplar replay \cite{34,36} stores real samples, and generative replay uses GANs to synthesize pseudo-samples. Recent discriminative selection strategies have diverged in their optimization objectives: some methods like GSS \cite{GSS} prioritize maximizing gradient variance to mitigate constraint conflicts, while others such as OCS \cite{OCS} and GCR \cite{GCR} focus on constructing compact representative coresets through multi-criteria streaming selection or weighted gradient approximation, respectively. Recent advances also focus on diffusion-based generative replay to improve sample quality: SDDGR \cite{33} integrates Stable Diffusion with pseudo-labeling to generate spatially consistent synthetic data, while the work in \cite{32} adopts cross-sampling strategies to bridge the domain gap between real and synthetic samples. However, these methods suffer from the high computational complexity of generative models and poor cross-modal consistency. Unlike them, our MPM module uses conformal prediction-based low-uncertainty sample replay to avoid generative model overhead while ensuring cross-modal alignment.

\section{Unified Framework for Sketch Biometric Identification}

\subsection{Problem Definition}
Sketch biometric identification aims to train a unified model capable of handling multiple arbitrary sketch-based biometric ReID tasks with a flexible training sequence. Let the entire dataset be denoted as \(D=\{D_1, D_2, ..., D_K\}\), where \(D_i\) represents the sub-dataset corresponding to the i-th sketch-based biometric ReID task. Each sub-dataset consists of a training set \(D_i^{tr}\) and a testing set \(D_i^{te}\). Samples in all datasets follow the core association of "identity-sketch set-photo set": each identity y corresponds to a set of sketch images \(X_y^{sketch}\), a set of photo images \(X_y^{photo}\), and a unique identity label y. All sub-datasets are completely independent. For the data availability constraint, the training process adopts a sequential task learning paradigm: when training on the training set of the i-th task \(D_i^{tr}\), where $i\geq 2$, the training data of all previously learned tasks are no longer fully accessible; only a subset of training samples can be used for each previously learned task. The testing set contains exclusively unseen real data. For evaluation, the model is required to separately conduct cross-modal matching on each test set \(D_i^{te}\) to complete the corresponding sketch-based biometric ReID task.

\begin{figure*}[t]
  \centering
  
   \includegraphics[width=0.7\textwidth]{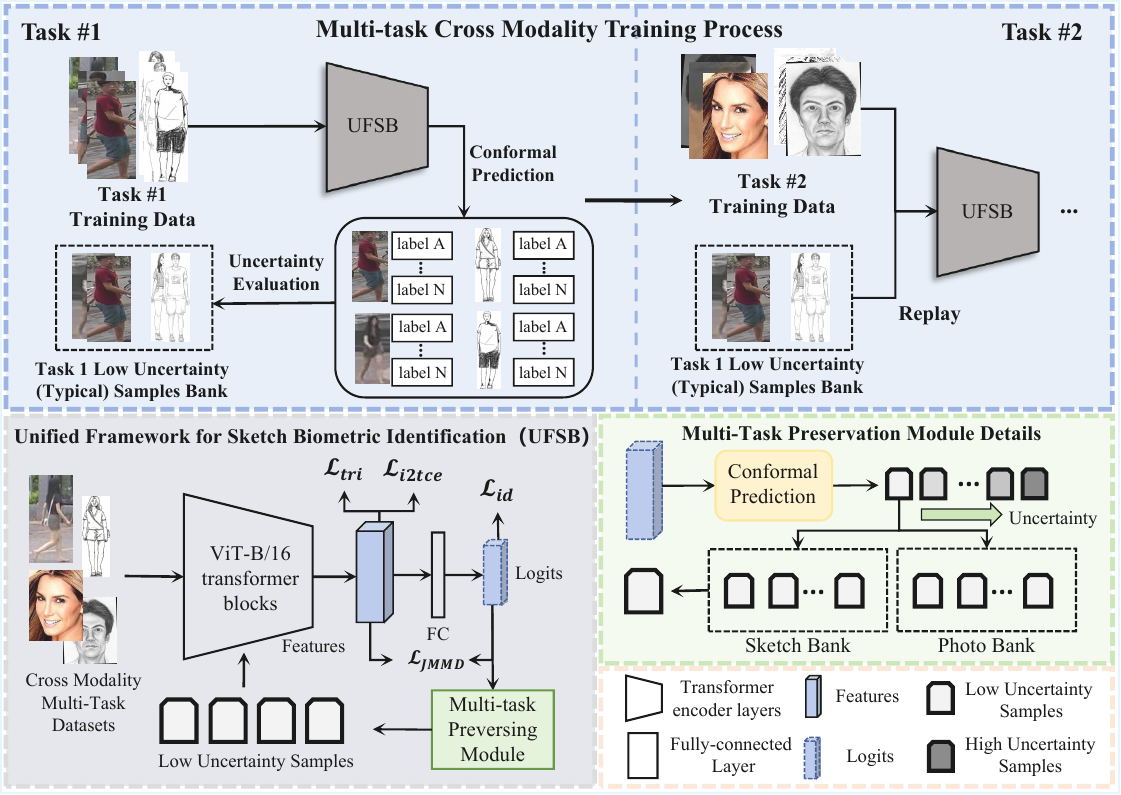}
    
   \caption{The overall pipeline of the Unified Framework for Sketch Biometric Identification (UFSB).}
   \label{fig:method}
\end{figure*}

\textit{The proposed pipeline of our method:}
As shown in Fig.\ref{fig:method}, the framework first trains on Task \#1, selects low-uncertainty typical samples via Conformal Prediction to store in sketch/photo banks, then rehearses these samples during Task  \#2 training to mitigate catastrophic forgetting. The core components include the ViT-B/16 transformer backbone for feature extraction, combined loss constraints, including Triplet loss \(\mathcal{L}_{tri}\), image-to-text cross-entropy loss \(\mathcal{L}_{i2tce}\), JMMD loss \(\mathcal{L}_{JMMD}\) for cross-modal alignment, and the Multi-Task Preservation Module (MPM) for sample selection and replay, enabling incremental accumulation of multi-task sketch biometric identification knowledge.

\subsection{Sketch Biometric Identification Foundation Module}
The SIM module serves as the core foundational component of the proposed unified sketch biometric identification framework, with its primary objective of establishing the model’s fundamental cross-modal re-identification capability. This module specifically addresses the modality gap inherent between sketch and photo images by extracting discriminative visual features and minimizing the distribution discrepancy between the two modalities, thereby laying a robust foundation of a unified feature space for subsequent multi-task learning.  

The SIM module’s core design centers on two pivotal components: the ClipReID backbone \cite{44clipreid} for feature extraction and the Joint Maximum Mean Discrepancy (JMMD) constraint \cite{43jan} for cross-modal distribution alignment. Built on CLIP’s ViT-B/16 architecture, the ClipReID backbone leverages CLIP’s pre-trained parameters with all layers fine-tuned to adapt to cross-modal ReID, ultimately extracting 768-dimensional global features. To align sketch and photo feature distributions, the JMMD constraint is enforced across multiple critical layers—acting on the 768-dimensional global features from the ClipReID backbone, the feature outputs of the penultimate transformer block, and the classification scores of both modalities.

The JMMD Loss is adopted to compute the distribution distance between sketch and photo feature sets, following the formulation in JAN \cite{43jan}. Specifically, let the sketch set S and photo set P consist of \(|S|\) and \(|P|\) samples, respectively. The deep encoder generates activations across multiple layers \(\mathcal{L}\) for sketch samples as \(\left\{ \left( z_{i,S}^1, ..., z_{i,S}^{|\mathcal{L}|} \right) \right\}_{i=1}^{|S|}\) and for photo samples as \(\left\{ \left( z_{i,P}^1, ..., z_{i,P}^{|\mathcal{L}|} \right) \right\}_{i=1}^{|P|}\), where \(z_{i,S}^l\) and \(z_{i,P}^l\) represent the activation of the i-th sample from layer \(l \in \mathcal{L}\) for sketch and photo, respectively. The JMMD between S and P is defined as the squared distance between their empirical kernel mean embeddings in the Reproducing Kernel Hilbert Space (RKHS), computed via multi-layer joint kernels:
\begin{equation}
    \begin{split}
    \mathcal{L}_{\text{JMMD}} &= \frac{1}{n_S^2} \sum_{i=1}^{n_S} \sum_{j=1}^{n_S} \prod_{l \in \mathcal{L}} k_l(z_{i,S}^l, z_{j,S}^l) \\
    &\quad + \frac{1}{n_P^2} \sum_{i=1}^{n_P} \sum_{j=1}^{n_P} \prod_{l \in \mathcal{L}} k_l(z_{i,P}^l, z_{j,P}^l) \\
    &\quad - \frac{2}{n_S n_P} \sum_{i=1}^{n_S} \sum_{j=1}^{n_P} \prod_{l \in \mathcal{L}} k_l(z_{i,S}^l, z_{j,P}^l),
    \end{split}
\end{equation}
where \(k_l(\cdot, \cdot)\) denotes the kernel function for layer l, and the product across \(l \in \mathcal{L}\) fuses multi-layer feature distributions to comprehensively capture cross-modal discrepancies.

To ensure the learning effectiveness of biometric re-identification, we follow the pipeline of ClipReID \cite{44clipreid} and adopt the same two-stage training paradigm as ClipReID during the training phase of each task dataset. Additionally, the classical ReID loss \(\mathcal{L}_{\text{ReID}}\) is employed, which consists of a Triplet loss \(\mathcal{L}_{\text{tri}}\), an ID loss \(\mathcal{L}_{\text{id}}\), and an image-to-text cross-entropy loss \(\mathcal{L}_{\text{i2tce}}\). \(\mathcal{L}_{\text{ReID}}\) is calculated as:
\begin{equation}
\mathcal{L}_{\text{ReID}} = \mathcal{L}_{\text{id}} + \mathcal{L}_{\text{tri}} + \mathcal{L}_{\text{i2tce}}.
\end{equation}

Ultimately, the losses utilized in SIM are summarized as follows:
\begin{equation}
\mathcal{L}_{\text{SIM}} = \mathcal{L}_{\text{ReID}} + \mathcal{\alpha} \mathcal{L}_{\text{JMMD}},
\end{equation}
where the hyperparameter \(\alpha\) is employed to balance the weights of the respective loss components. The specific setting of \(\alpha\) is detailed in Section \ref{sec:Experiment}.

\subsection{Multi-Task Preservation Module}

In this section, we aim to enable the model to retain the discriminative knowledge of the previous task while avoiding interference with the learning of the current new task. This is the core objective of the Multi-Task Preservation Module (MPM) operates in parallel with the SIM without a primary-secondary hierarchy, as illustrated in Fig. \ref{fig:method}. Specifically, for each sample x from the previous task, the SIM first extracts its feature and outputs a class probability distribution:
\begin{equation}
    \boldsymbol{\pi}(x) = [\pi_1, \pi_2, ..., \pi_C],
\end{equation}
where C denotes the number of identities in the previous task, and \(\pi_y \geq 0\) with the sum of all \(\pi_y\) equal to 1. 

Based on this probability distribution, MPM constructs a Conformal Prediction set \(C(x)\) to ensure the true label of x is covered with a high finite-sample confidence while minimizing the size of \(C(x)\) to avoid redundant information, thus guaranteeing the reliability of subsequent sample selection. To quantify the uncertainty of each sample, we define the uncertainty score as:
\begin{equation}
    \text{Unc}(x) = |C(x)| + \text{Conf}(x),
\end{equation}
where \(|C(x)|\) is the size of the prediction set and
\begin{equation}
    \text{Conf}(x) = \max_{y \in C(x)} \pi_y - \min_{y \in C(x)} \pi_y,
\end{equation}
is a smaller value reflects more consistent confidence in the identities within \(C(x)\). MPM then maintains two dedicated repositories, a sketch bank and a photo bank, to store low-uncertainty samples from the previous task: for each identity in the previous task, only one sketch sample and one photo sample are retained, and when the storage capacity for an identity is reached, existing samples are replaced by new ones with lower uncertainty to ensure stored samples always represent the most reliable discriminative knowledge. During the training of the current new task, we adopts an epoch-alternating replay strategy: it alternates between one epoch of training on new task data and one epoch of training on stored low-uncertainty samples, ensuring the model balances the acquisition of new task knowledge and the retention of previous task knowledge.

To formalize the core logic of MPM, we first define key variables aligned with SIM’s notation: let \(o_y\) be the rank of identity y sorted by \(\pi_y\) in descending order, and 
\begin{equation}
    \rho_y = \sum_{\substack{y' \in \mathcal{Y} \\ o_{y'} < o_y}} \pi_{y'},
\end{equation}
where \(\mathcal{Y} = \{1, 2, ..., C\}\) is the identity space of the previous task, and \(\rho_y\) represents the cumulative probability of all identities with a higher rank than y. For each identity y, we calculate a CP score to determine whether it is included in \(C(x)\):
\begin{equation}
    \text{score}_y = \rho_y + \pi_y + \lambda \cdot \max(0, o_y - k_{\text{reg}}),
\end{equation}
where \(\lambda = 0.3\) penalizes low-rank identities to prevent excessive expansion of \(C(x)\), and \(k_{\text{reg}} = 10\) defines the rank beyond which penalties are applied. The prediction set \(C(x)\) is then constructed by including all identities with \(\text{score}_y \leq \tau\) ,where \(\tau = 5.0\) is a confidence threshold:
\begin{equation}
    C(x) = \left\{ y \in \mathcal{Y} \mid \text{score}_y \leq \tau \right\}.
\end{equation}

MPM preserves the cross-modal alignment capability while retaining previous task knowledge. Compared to traditional Experience Replay methods, MPM’s CP-based sample selection guarantees the reliability of stored samples. Moreover, replay epochs prevent the model’s parameters from drifting away from the parameter space corresponding to the discriminative features of the previous task, thus mitigating catastrophic forgetting; at the same time, the shared cross-modal knowledge indirectly enhances the performance of the new task, which is verified by experimental results in Sec. \ref{sec:Experiment}.

\section{Benchmark and Evaluation Protocol}

\subsection{A New Unified Sketch Biometric Identification Benchmark}
To facilitate the research on unified cross-modal sketch biometric identification, we propose a large-scale, task-unified benchmark, SketchUnified-BioID. This benchmark addresses the limitations of existing single-task sketch datasets and provides a standardized evaluation platform for multi-task lifelong sketch biometric learning.
More detailed information about the SketchUnified-BioID benchmark can be found in the \emph{supplementary material}.

\textbf{SketchUnified-BioID} features a hierarchical structure tailored for cross-task and cross-modal evaluation. It is partitioned by task type into two core subsets: the Sketch-Pedestrian ReID Subset and the Sketch-Face ReID Subset. Each subset is further divided by data authenticity into a real data partition and a synthetic data partition. The real data partition is dedicated to standard training and evaluation, while the synthetic data partition serves exclusively for auxiliary training to alleviate the scarcity of real sketch data. A critical design constraint underpins the benchmark: identities are fully non-overlapping across all sub-datasets, and training and test identities within each task remain strictly disjoint. This ensures objective evaluation of model generalization without data leakage.

\textit{Sketch-Pedestrian ReID Subset.} This subset serves as the evaluation platform for sketch-pedestrian matching. Its real data partition adopts the public MaSk1K dataset, retaining its original train-test split. The synthetic data partition comprises our self-constructed CUHK03-Sketch dataset, which is used solely for auxiliary training.
\textit{Sketch-Face ReID Subset.} Targeting sketch-face matching evaluation, this subset’s real data partition integrates two public datasets, CUFSF and IIIT-D Viewed Sketch, while preserving their original train-test splits. The synthetic data partition employs our self-built CelebA-Sketch dataset, designed exclusively for auxiliary training.

\textit{Remarks:} compared with existing sketch-related benchmarks, SketchUnified-BioID has three core innovations: 1) Cross-Task Unification: It is the first benchmark to unify sketch-pedestrian ReID and sketch-face ReID tasks, while existing benchmarks only focus on a single task. 2) Large Scale in Identities and Samples: The benchmark covers 13,942 identities, far exceeding existing sketch benchmarks in both identity count and sample size. 3) Integration of Real and Synthetic Data: It fuses real data and large-scale synthetic data, addressing the scarcity of real sketch data.

\subsection{Detailed Sketch-Pedestrian ReID Subset}
\textit{Real Data Partition.}
MaSk1K (Market-Sketch-1K) is a large-scale public sketch-pedestrian ReID dataset with multi-perspective and multi-style characteristics, which we adopt as the real data partition for sketch-pedestrian evaluation. It contains 4,763 sketches of 996 identities and 32,668 photos of 1,501 identities. A key feature of MaSk1K is that each identity has multiple sketches, reflecting both witnesses’ subjective multi-perspective cognition of the same individual and style variations from different artists, which mimics real-world forensic scenarios. We strictly follow the original train-test split of MaSk1K to ensure consistency with existing evaluations.

\textit{Synthetic Data Partition.}
To supplement the scarce real sketch data, we construct the CUHK03-Sketch dataset via our proposed fast synthesis pipeline, using the CUHK03-NP dataset as the source. CUHK03-Sketch contains 28,194 images of 1,467 identities, with a 1:1 ratio of synthetic sketches to photos. The original CUHK03 dataset includes 14,097 images of 1,467 identities, collected by 6 campus cameras, ensuring diverse pose and viewpoint variations. CUHK03-Sketch is exclusively used for auxiliary training to enhance model generalization and is not involved in any evaluation.

\subsection{Detailed Sketch-Face ReID Subset}

\textit{Real Data Partition.}
The real data partition for sketch-face ReID integrates two public datasets, retaining their original train-test splits to maintain evaluation fairness:

CUFSF (CUHK Face Sketch FERET Database): Focuses on face sketch synthesis and recognition, containing 1,194 identities. Each identity has 1 photo with lighting variations and 1 sketch with shape exaggeration drawn by a professional artist, resulting in 1,194 photo-sketch pairs. CUFSF is widely used as a standard benchmark for face sketch recognition, making it suitable for baseline evaluation.

IIIT-D Viewed Sketch: A challenging sketch-face dataset consisting of 238 photo-sketch pairs. The photos are sourced from three databases: 67 pairs from the FG-NET aging database, 99 pairs from the Labeled Faces in the Wild (LFW) database, and 72 pairs from the IIIT-D student staff database. All sketches are hand-drawn by a professional artist, introducing natural style variations and pose differences—making it ideal for evaluating model robustness to real-world sketch variations.

\textit{Synthetic Data Partition.}
The proposed unreal face sketch dataset is called the CelebA-Sketch Dataset. These face images in the CelebA dataset \cite{40} cover large pose variations and background clutter, which more representative of the real-world scenes. To promote the heterogeneous face analysis field and evaluate the generalization ability, we build the corresponding face sketches to face photos in the CelebA database \cite{40}. In the preprocessing stage, we directly utilized the face detection method \cite{41} to align and crop images from original photos, which follows the same strategy in \cite{42}. To simulate the fingerstyle of artists in hand-drawn sketches, we utilize image processing 2 to generate pencil sketches. From this perspective, they are similar to line drawing sketches. Different from the minimal facial sketches by plotting landmarks, these generated line drawing sketches are easier to distinguish and fit the real-world scenario. The introduced large-scale CelebA-sketch database contains 200k photo-sketch pairs belonging to 10k identities. Abundant extra labels, like landmarks and attributes, would help inspire other researchers. Noting that these sketches contain less texture information and a higher abstract level compared with hand-drawn sketches.

\subsection{In-depth Explanation of Core Innovations}

\textit{Cross-Task Unification.}
Existing sketch benchmarks are limited to single-task evaluation, failing to support unified biometric identification across multiple sketch-based tasks. SketchUnified-BioID is the first benchmark to unify sketch-pedestrian and sketch-face ReID, with a scalable structure that can be extended to other sketch biometrics. It fills the gap between single-task sketch ReID and multi-task lifelong learning.

\textit{Large-Scale Coverage.}
Most existing sketch datasets have limited identity and sample scales. SketchUnified-BioID covers 13,942 identities, with a total of over 480k images. This large scale ensures sufficient data diversity for training and evaluating multi-task models.

\textit{Real-Synthetic Synergy.}
Real sketch data suffers from high annotation costs, limited scale, and privacy risks. Existing benchmarks rely solely on real data, leading to poor model generalization. SketchUnified-BioID fuses real data and large-scale synthetic data which not only reduces data collection costs and avoids privacy risks but also supplements edge cases. This effectively addresses the scarcity of real sketch data.

\subsection{Evaluation Protocols and Metrics}
To comprehensively evaluate the model performance, we propose two training and evaluation schemes for SketchUnified-BioID:
\textbf{Scheme A}: The MaSk1K dataset is used as the real training and test dataset for the sketch re-identification task, while the CUFSF dataset serves as the real training and test dataset for the face sketch recognition task. Additionally, the CelebA-Sketch and CUHK03-Sketch datasets can be employed as non-real auxiliary training datasets for sketch re-identification and face sketch recognition, respectively, without participating in testing. Training is conducted in two-step incremental learning, and the model's performance on both tasks is evaluated simultaneously at each assessment step.
\textbf{Scheme B:} Built upon Scheme A, this scheme replaces the real training and test dataset for the face sketch recognition task with the more challenging IIIT-D Viewed Sketch dataset, while keeping all other configurations unchanged.
By calculating the mean Average Precision (mAP) and Rank@1 accuracy (R@1) on each dataset, the model's performance in specific domains is evaluated. Meanwhile, the average mAP and average R@1 across all tasks are statistically analyzed to compare the model's comprehensive sketch biometric identification performance.

\section{Experimental Results}
\label{sec:Experiment}
\subsection{Implementation Details}
Our Unified Framework for Sketch Biometric Identification model (UFSB) training employs the Adam optimizer, with a 10-epoch warm-up. The learning rate linearly increases from $(5 \times 10)$ to $(5 \times 10)$ during warm-up, followed by $0.1 \times$ decay at epochs 30 and 50. The first task is trained for 60 epochs. The second task is trained for 30 epochs. For these experiments, we set the weight hyperparameter \(\alpha\) of the JMMD Loss to 5.0. Each mini-batch contains 64 images $(B= 16\times 4)$, including 16 identities with 4 samples per identity. Input images are resized to $(256 \times 128)$ with random horizontal flipping, padding, cropping, and erasing augmentation. All experiments run on a single NVIDIA 3090 GPU.
\subsection{The Compared Methods}
We compare our method against three types of representative methods. First, cross-modal ReID methods with BDG \cite{10BDG} as the state-of-the-art baseline. Second, lifelong ReID methods including LwF \cite{45lwf}, DKP \cite{22dkp} and DASK \cite{23dask}; LwF adopts the same backbone and experimental settings as our method, while DKP and DASK are non-exemplar methods implemented via their official code. Third, the JointTrain method, a widely recognized upper bound for lifelong ReID that uses all available training data simultaneously. To ensure a fair and rigorous comparison, the Joint Training baseline strictly adheres to the identical experimental configuration as our UFSB framework. Specifically, it employs the same ClipReID backbone, the same combined loss function, and the same optimization settings. Tables 1, 2, and 3 present the average performance of different methods, respectively. Among the results, the optimal performance is marked in red, while the second-best performance is marked in blue.

\subsection{Comparison Results}
\begin{table}[t]
  \caption{Comparison of different methods on MaSk1K datasets. $^{+}$ denotes that the proposed synthetic datasets CelebA-Sketch and CUHK03-Sketch are used for auxiliary training.}
  \label{tab:method_comparison}
  \centering
  \begin{tabular}{@{}lcccc@{}}  
    \toprule  
    Method & mAP & rank@1 & rank@5 & rank@10 \\
    \midrule  
    DDAG \cite{DDAG} & 12.13 & 11.22 & 25.40 & 35.02 \\
    CM-NAS \cite{CMNAS} & 0.82 & 0.70 & 2.00 & 3.90 \\
    CAJ \cite{CAJ} & 2.38 & 1.48 & 3.97 & 7.34 \\
    MMN \cite{MMN} & 10.41 & 9.32 & 21.98 & 29.58 \\
    DART \cite{DART} & 7.77 & 6.58 & 16.75 & 23.42 \\
    DCLNet \cite{DCLNet} & 13.45 & 12.24 & 29.20 & 39.58 \\
    DSCNet \cite{DSCNet} & 14.73 & 13.84 & 30.55 & 40.34 \\
    DEEN \cite{DEEN} & 12.62 & 12.11 & 25.44 & 30.94 \\
    BDG \cite{10BDG} & 19.61 & 18.10 & 38.95 & 57.75 \\
    \midrule 
    UFSB(Ours) & \underline{27.08} & \underline{32.11} & \underline{53.50} & \underline{64.35} \\
    \(\mathrm{UFSB(Ours)}^{+}\) & \textbf{28.25} & \textbf{34.51} & \textbf{55.36} & \textbf{66.16} \\
    \bottomrule  
  \end{tabular}
\end{table}
\textit{Sketch-Pedestrian ReID Performance on MaSk1K}: among cross-modal ReID methods, BDG stands as the state-of-the-art baseline with 19.61\% mAP and 18.10\% R@1. It still suffers from poor generalization to diverse drawing styles due to scarce real sketch data. Our method achieves a significant breakthrough with \textbf{28.25\% mAP} and \textbf{34.51\% R@1}, outperforming BDG by 8.64 percentage points in mAP and 16.40 percentage points in R@1. This superiority comes from two core designs. The SIM module uses Joint Maximum Mean Discrepancy to explicitly align sketch and photo feature distributions, resolving the modality gap. Fusing large-scale synthetic sketches supplements scarce real data, enabling the model to generalize across diverse sketch drawing styles, an ability unavailable in methods trained solely on real samples. These results validate that our framework effectively addresses the dual challenges of cross-modal mismatch and style diversity in sketch-pedestrian ReID, establishing a strong foundation for unified sketch-based biometric identification across pedestrian and face tasks.

\begin{table}[t]
  \caption{Comparison results on Scheme A. Scheme A adopts MaSk1K for the sketch ReID task and CUFSF for the sketch-face recognition task.}
  \label{tab:dataset_compare_A}
  
  \centering
  \resizebox{0.98\linewidth}{!}{
  \begin{tabular}{@{}lcccccc@{}}  
    \toprule
    Method & \multicolumn{2}{c}{MaSk1K} & \multicolumn{2}{c}{CUFSF} & \multicolumn{2}{c}{Avg} \\
    \cmidrule(lr){2-3} \cmidrule(lr){4-5} \cmidrule(lr){6-7}  
    & mAP & R@1 & mAP & R@1 & mAP & R@1 \\
    \midrule
    \(\mathrm{Joint}^{+}\) & \underline{$26.78$} & \underline{$30.59$} & \textbf{98.62} & \textbf{97.55} & \textbf{62.7} & \textbf{64.07} \\
    \(\mathrm{LWF}^{+}\) \cite{45lwf} & 5.96 & 7.38 & 84.63 & 78.24 & 45.30 & 42.81 \\
    DKP \cite{22dkp} & 9.8 & 11.1 & 84.7 & 78.1 & 47.3 & 44.6 \\
    DASK \cite{23dask} & 0.5 & 0.3 & 61.2 & 54.3 & 30.8 & 27.3 \\
    \(\mathrm{UFSB(Ours)}^{+}\) & \textbf{27.19} & \textbf{31.69} & \underline{$96.59$} & \underline{$94.24$} & \underline{$61.89$} & \underline{$62.97$} \\
    \bottomrule
  \end{tabular}
}
\end{table}

\begin{table}[t]
  \caption{Comparison results on Scheme B. Scheme B adopts MaSk1K for the sketch ReID task and IIIT-D Viewed Sketch for the sketch-face recognition task.}
  \label{tab:dataset_compare_B}
  \centering
  \resizebox{0.98\linewidth}{!}{
  \begin{tabular}{@{}lcccccc@{}}  
    \toprule
    Method & \multicolumn{2}{c}{MaSk1K} & \multicolumn{2}{c}{IIIT-D Sketch} & \multicolumn{2}{c}{Avg} \\
    \cmidrule(lr){2-3} \cmidrule(lr){4-5} \cmidrule(lr){6-7}  
    & mAP & R@1 & mAP & R@1 & mAP & R@1 \\
    \midrule
    \(\mathrm{Joint}^{+}\) & \textbf{26.83} & \textbf{30.21} & \underline{83.8} & \underline{80.0} & \underline{55.32} & \underline{55.11} \\
    \(\mathrm{LWF}^{+}\) \cite{45lwf} & 5.69 & 5.91 & 53.87 & 42.86 & 29.78 & 24.39 \\
    DKP \cite{22dkp} & 5.2 & 6.2 & 57.2 & 42.1 & 31.2 & 24.2 \\
    DASK \cite{23dask} & 0.5 & 0.5 & 12.6 & 5.0 & 6.5 & 2.7 \\
    \(\mathrm{UFSB(Ours)}^{+}\) & \underline{25.19} & \underline{29.54} & \textbf{88.26} & \textbf{85.71} & \textbf{56.73} & \textbf{57.63} \\
    \bottomrule
  \end{tabular}
}
\end{table}

\begin{figure}[t]
  \centering
  
   \includegraphics[width=0.98\linewidth]{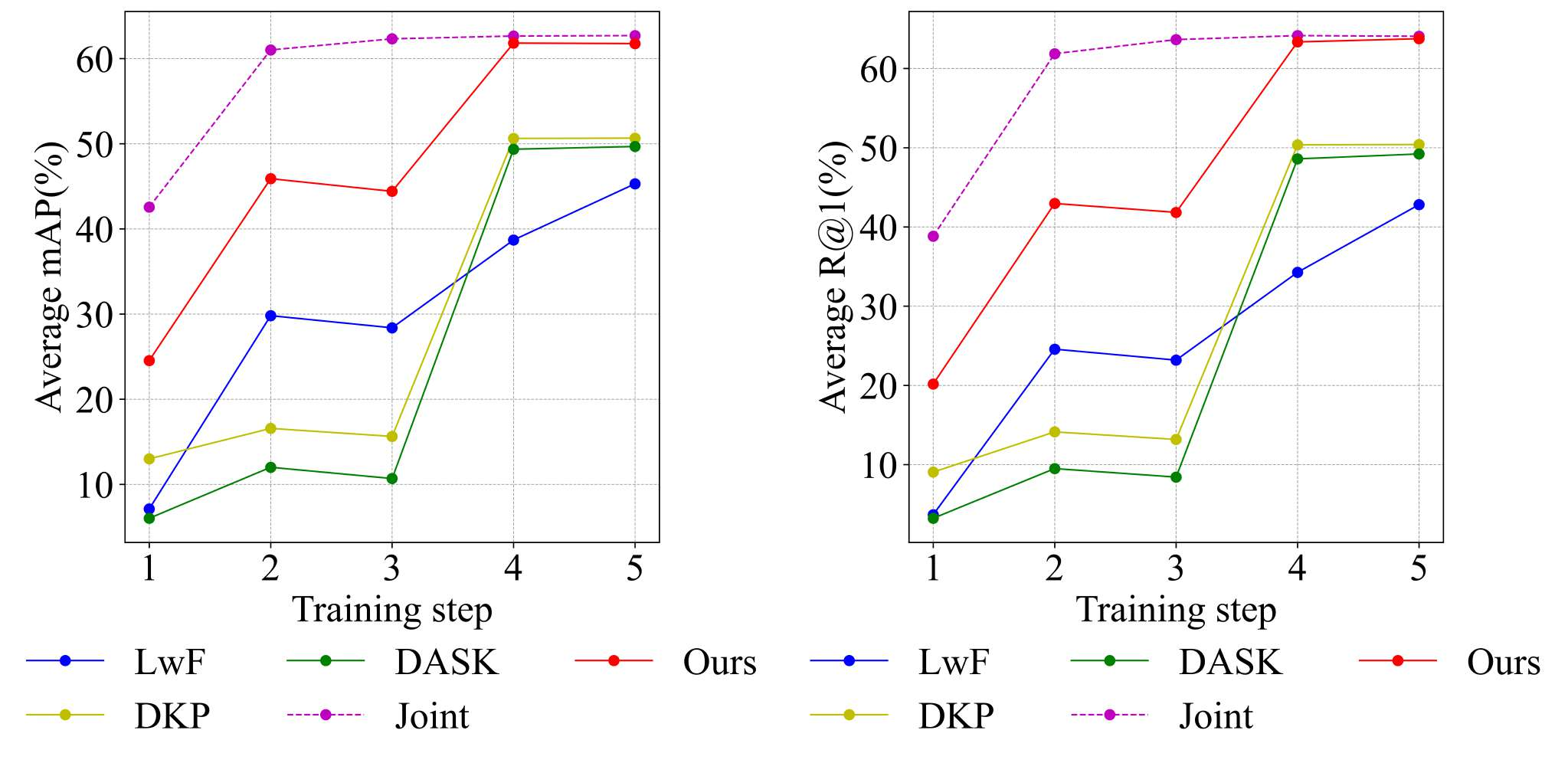}
    
   \caption{Performance tendency on Scheme A. Training steps 1–5 correspond to the start of training, halfway through first task training, completion of first task training, halfway through second task training, and completion of second task training, respectively.}
   \label{fig:cufsf}
\end{figure}

\begin{figure}[t]
  \centering
  
   \includegraphics[width=0.98\linewidth]{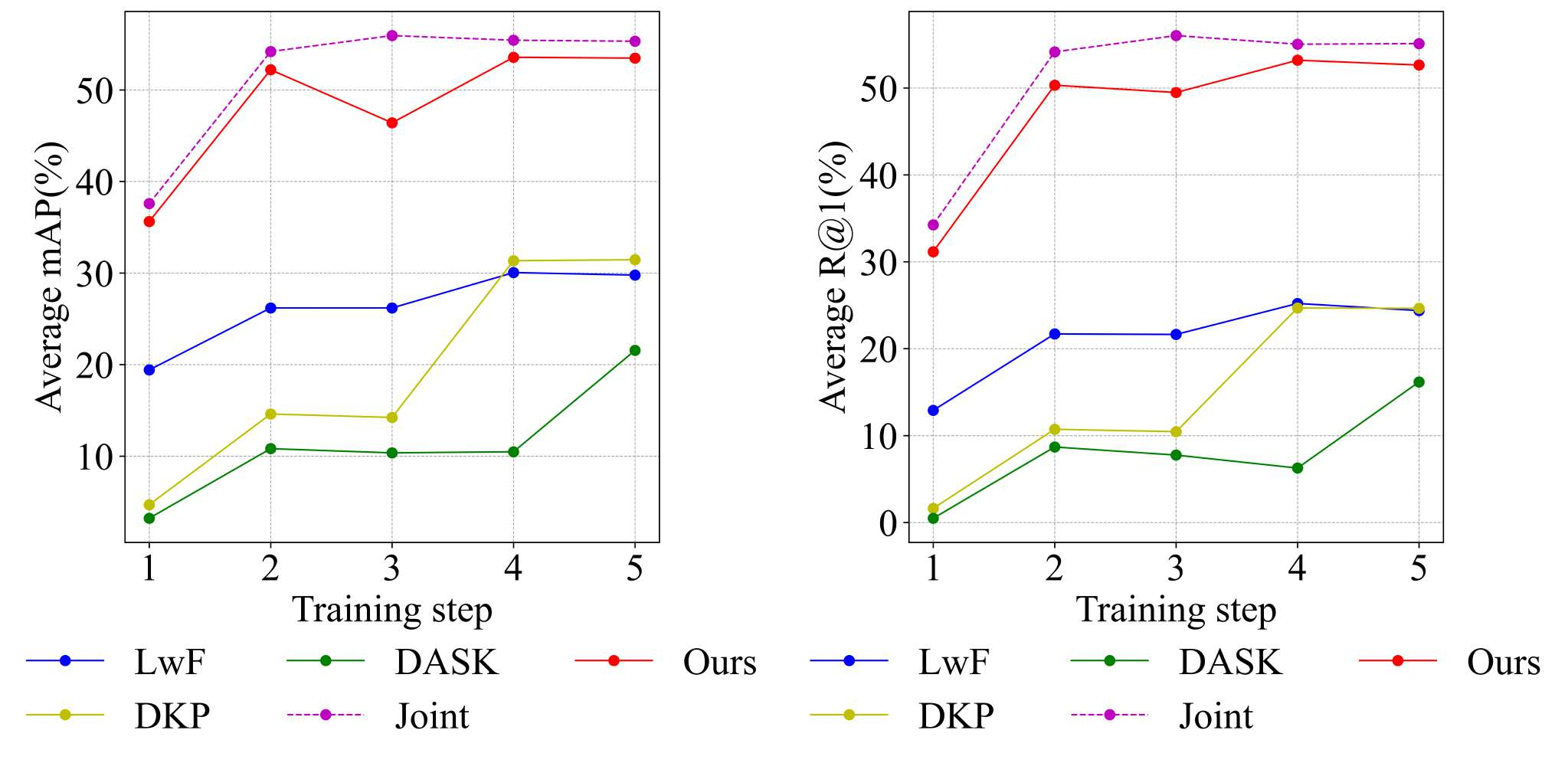}
    
   \caption{Performance tendency on Scheme B.}
   \label{fig:iiit}
\end{figure}

\textit{Multi-task Sketch Biometric Identification Performance under Scheme A and Scheme B.} 
Scheme A adopts CUFSF for the initial sketch-face ReID task and MaSk1K for the subsequent sketch-pedestrian ReID task, while Scheme B replaces CUFSF with the more challenging IIIT-D Viewed Sketch to evaluate generalization in realistic, variable scenarios. Quantitative results are summarized in Tables \ref{tab:dataset_compare_A} and \ref{tab:dataset_compare_B}, with detailed performance tendencies visualized in Figs. \ref{fig:cufsf} and \ref{fig:iiit}. Notably, to verify robustness against task ordering, we also conducted a \textit{Reverse Scheme A} (MaSk1K $\to$ CUFSF), where the training sequence is inverted; the corresponding results are presented in Table \ref{tab:dataset_compare_A_reverse}.

Lifelong ReID methods share a common limitation: they achieve relatively high performance on the new face task but struggle drastically with the previously learned pedestrian task, rendering them unsuitable for unified biometric identification. This imbalance stems from their inability to handle cross-task continual learning. Lifelong ReID methods fail to retain discriminative knowledge of the pedestrian task during face task training, leading to severe catastrophic forgetting. As a result, their pedestrian task performance drops to extremely low levels, while their face task performance remains acceptable

Our method achieves outstanding and balanced performance across both tasks under both schemes, thanks to the Multi-Task Preservation Module. Under Scheme A, it reaches 27.19\% mAP and 31.69\% R@1 on MaSk1K, and 96.59\% mAP and 94.24\% R@1 on CUFSF. While slightly lower than DASK’s face task performance, it achieves far more balanced performance across tasks. Under Scheme B, it achieves 25.19\% mAP and 29.54\% R@1 on MaSk1K, and 88.26\% mAP and 85.71\% R@1 on IIIT-D Viewed Sketch, demonstrating strong adaptability to challenging data. Compared to lifelong ReID methods, our method outperforms them by a large margin: under Scheme A, average mAP and R@1 surpass LwF by 16.59 and 20.16 percentage points, DKP by 11.23 and 12.57 percentage points, and DASK by 12.22 and 13.76 percentage points; under Scheme B, average mAP is 25.16 to 35.18 percentage points higher than these methods. The upper-bound Joint method, which uses all training data simultaneously, reaches 55.32\% average mAP and 55.11\% average R@1 under Scheme B, while our UFSB reaches 56.73\% and 57.63\%. While Joint exhibits marginal advantages on the MaSk1K dataset, UFSB delivers a significant performance boost on IIIT-D Sketch.
\emph{This consistent improvement in average metrics indicates that our UFSB effectively captures more intrinsic, task-shared discriminative features across sketch-based biometric identification tasks, thus enabling robust generalization across diverse sketch biometric scenarios.}

Performance tendencies in Figs.\ref{fig:cufsf} and Figs.\ref{fig:iiit} further validate our framework’s stability, anti-forgetting capability, and adaptive learning ability. Unlike lifelong ReID methods that suffer from catastrophic forgetting, our method maintains consistent or even improved performance across both tasks. After completing training on CUFSF or IIIT-D Viewed Sketch, its performance on MaSk1K remains stable, and its performance on the new task continues to improve. This stability and adaptability originate from the MPM module’s trusted sample replay, which not only preserves pedestrian-specific discriminative knowledge without interfering with new task learning but also enables synergistic learning between pedestrian and face tasks. More detailed comparison plots of performance trends between our method and other LReID methods under the two schemes are presented in Sec. \ref{subsec:vis} \textit{Visualization and Trend Analysis.}

\subsection{Visualization and Trend Analysis}
\label{subsec:vis}

\begin{figure}[t]
  \centering
  
   \includegraphics[width=\linewidth]{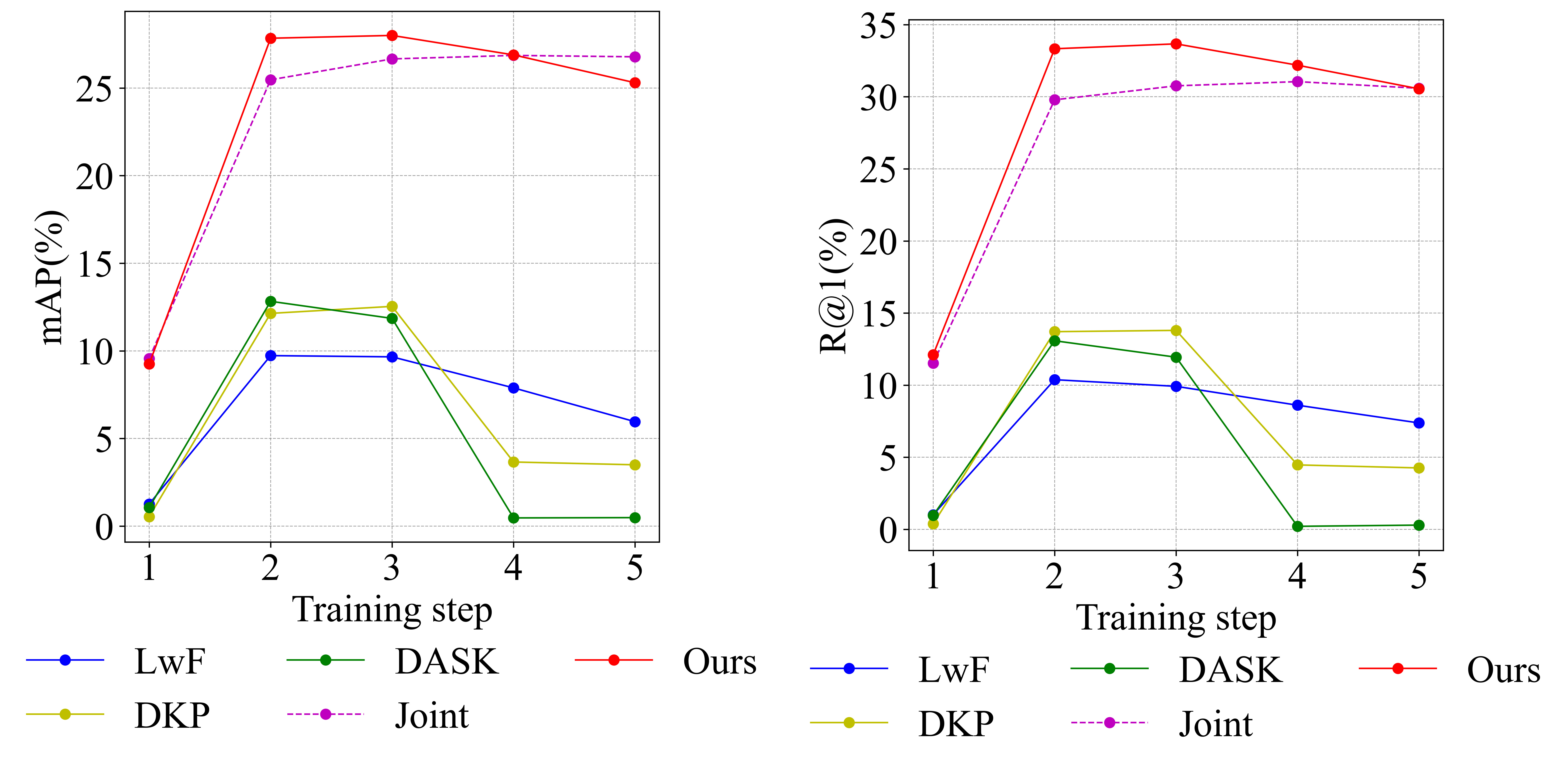}
    
   \caption{Performance tendency on the MaSk1k dataset of Scheme A (MaSk1k $\to$ CUFSF). Training steps 1–5 correspond to the start of training, halfway through the first task training, completion of first task training, halfway through the second task training, and completion of second task training, respectively.}
   \label{fig:A_ms1k}
\end{figure}

\begin{figure}[t]
  \centering
  
   \includegraphics[width=\linewidth]{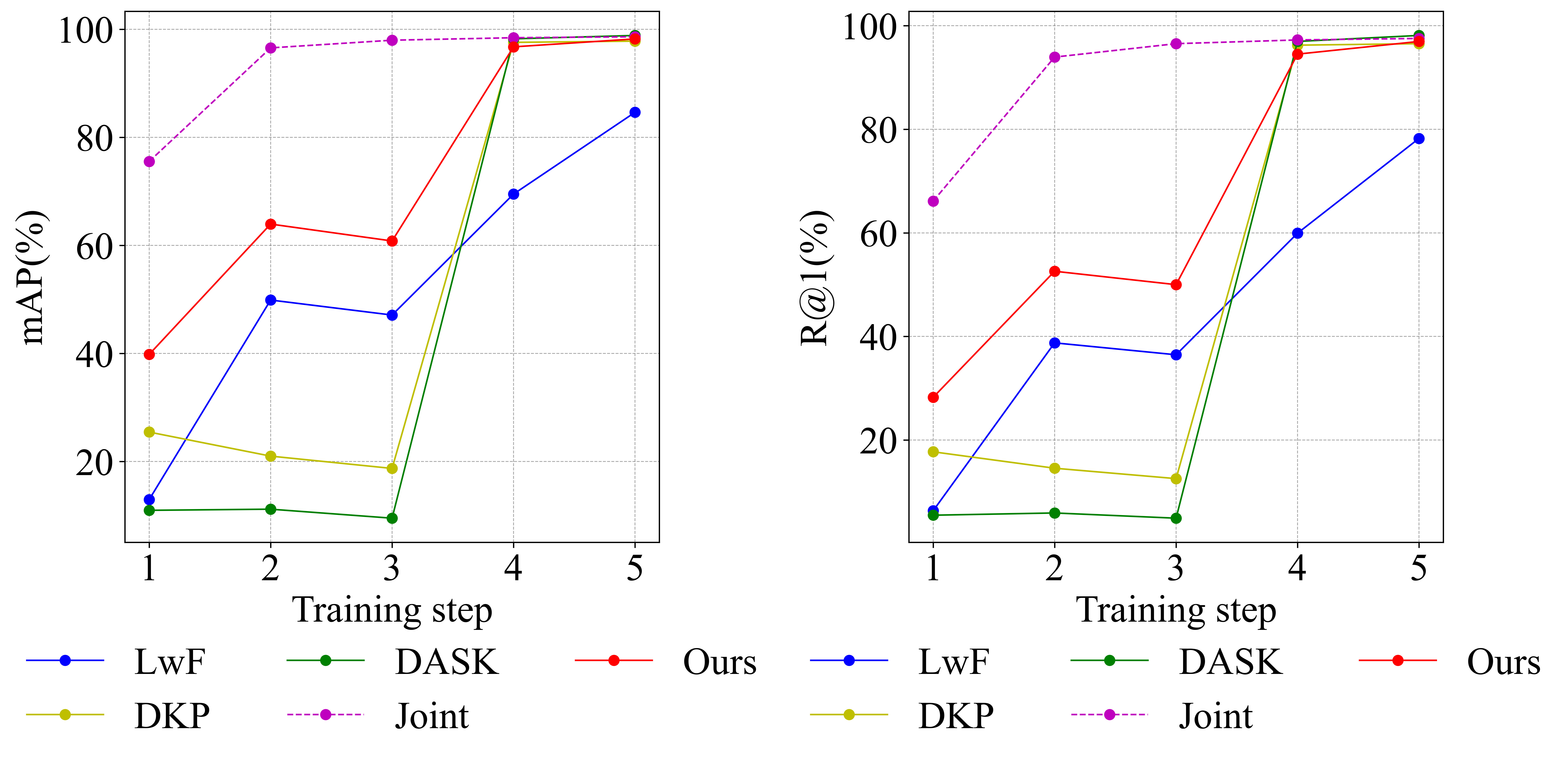}
    
   \caption{Performance tendency on the CUFSF dataset of Scheme A (MaSk1k $\to$ CUFSF).}
   \label{fig:A_cufsf}
\end{figure}

\begin{figure}[t]
  \centering
  
   \includegraphics[width=\linewidth]{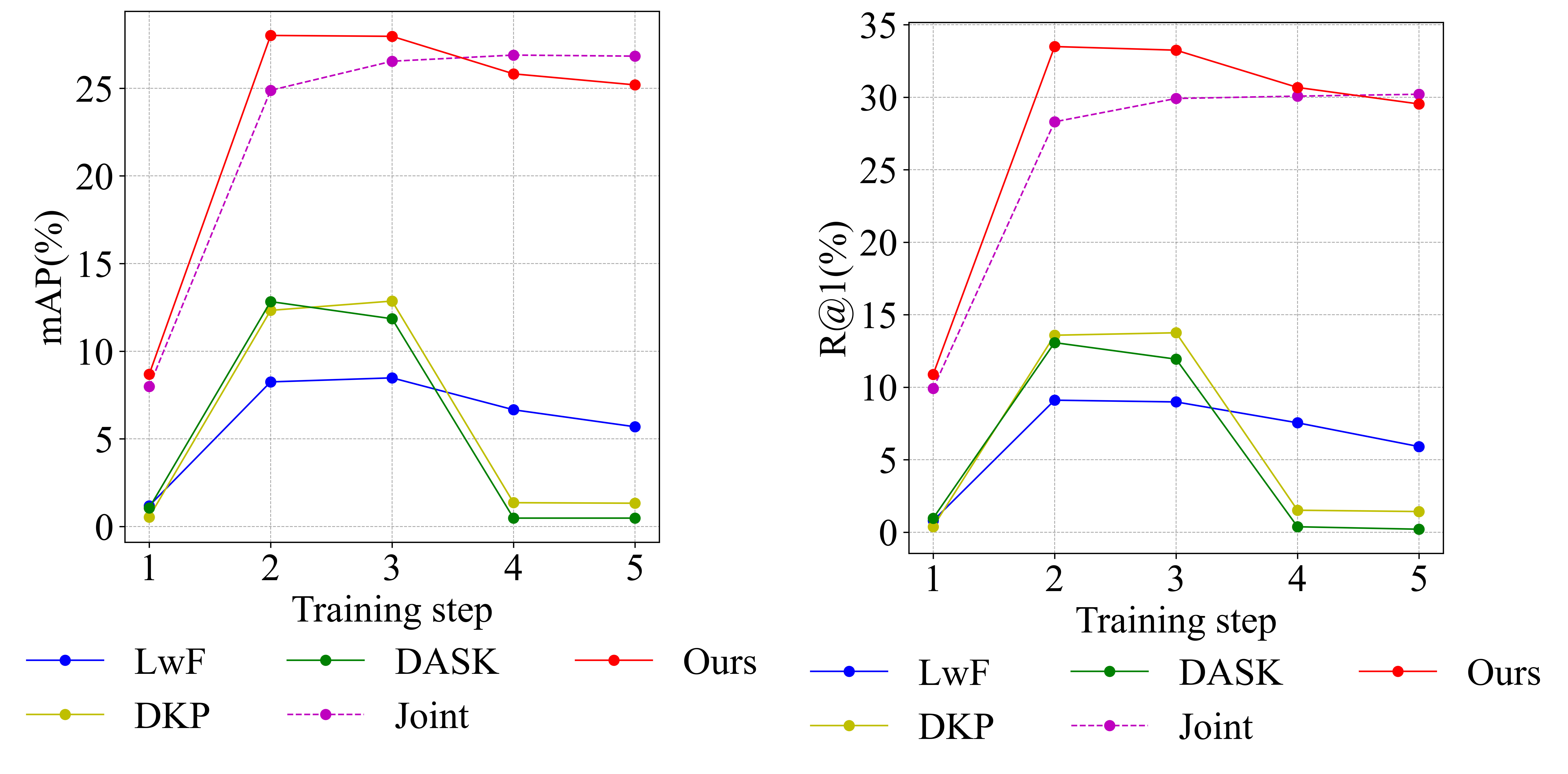}
    
   \caption{Performance tendency on the MaSk1k dataset of Scheme B (MaSk1k $\to$ IIIT-D Viewed Sketch).}
   \label{fig:B_ms1k}
\end{figure}

\begin{figure}[t]
  \centering
  
   \includegraphics[width=\linewidth]{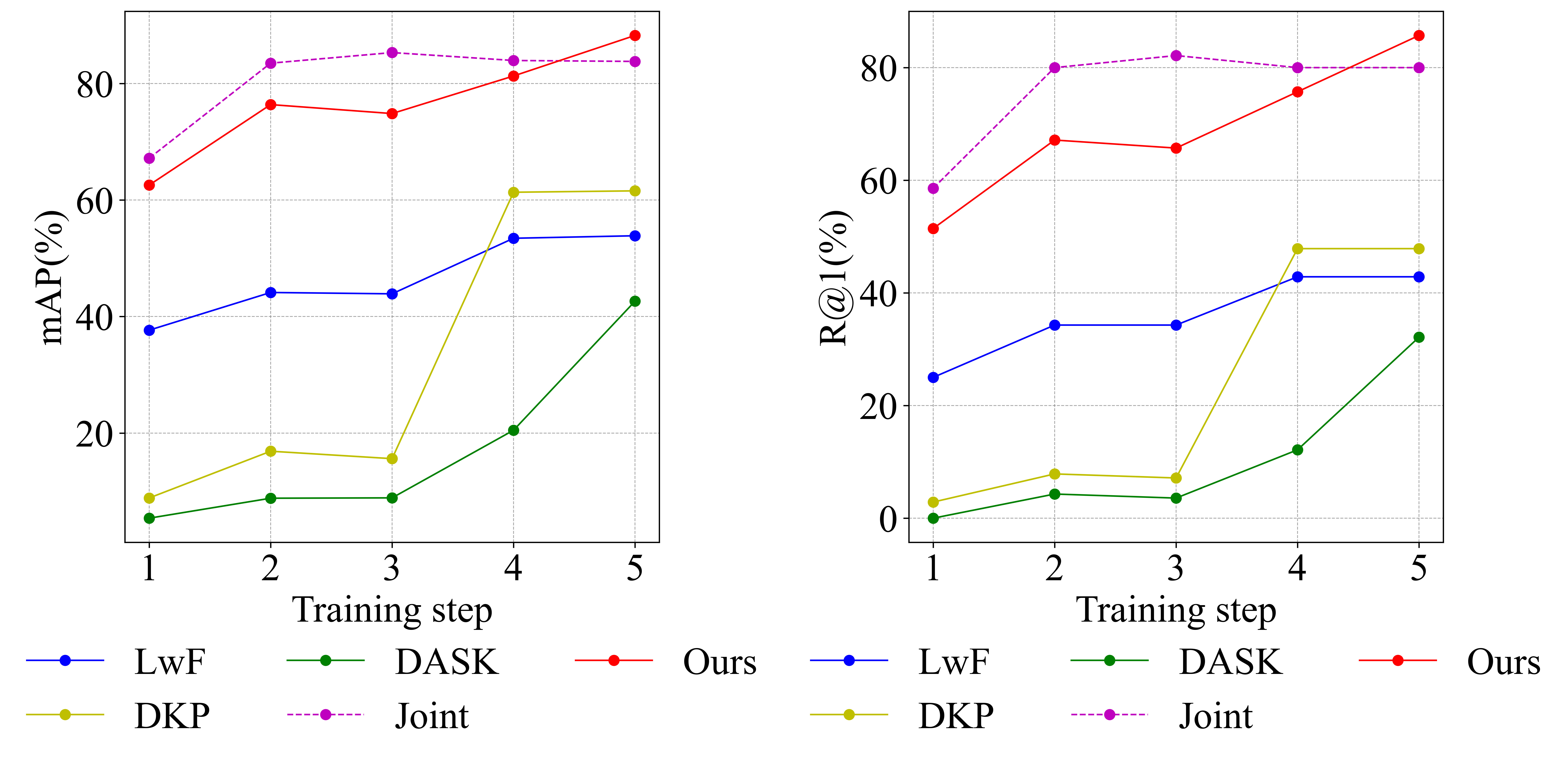}
    
   \caption{Performance tendency on the IIIT-D Viewed Sketch dataset of Scheme B (MaSk1k $\to$ IIIT-D Viewed Sketch).}
   \label{fig:B_iiit}
\end{figure}

To gain deeper insights into the learning dynamics and catastrophic forgetting behaviors, we visualize the performance trends of all methods across training steps under Scheme A and Scheme B. These trends reveal how different methods balance plasticity and stability.

\textit{Performance Trends under Scheme A.}
Scheme A adopts MaSk1K for sketch-pedestrian ReID and CUFSF for sketch-face ReID. The performance trajectories on these two datasets are illustrated in Figs. \ref{fig:A_ms1k} and \ref{fig:A_cufsf}, respectively.

Fig. \ref{fig:A_ms1k} depicts the performance variation on the MaSk1K dataset. At the completion of the first task (Training Step 3), our method achieves the highest mAP and R@1 among all continual learning baselines, establishing a robust foundation for pedestrian ReID. However, as training proceeds to the second task (Steps 4--5), existing lifelong ReID methods (LwF, DKP, DASK) suffer from severe degradation. Notably, LwF's mAP drops by over $70\%$ from its peak, while DKP and DASK nearly lose their discriminative capability (R@1 $\approx$ 0\%). This drastic decline confirms their inability to retain pedestrian features when exposed to face biometric data. In sharp contrast, our UFSB framework maintains remarkable stability throughout the second phase. Its metrics exhibit only minor fluctuations, staying at the leading level. This validates the efficacy of our MPM module: by replaying low-uncertainty samples, it effectively prevents parameter drift and preserves the previously learned sketch-pedestrian representations.

Fig. \ref{fig:A_cufsf} presents the trends on the CUFSF dataset. While all methods start from zero and improve during the second task, a clear trade-off emerges for baselines: DASK and DKP achieve decent final performance on the new task but at the cost of completely forgetting the old one. Our method, however, achieves a superior balance. Its final mAP and R@1 on CUFSF are comparable to state-of-the-art baselines, indicating that the MPM's replay mechanism does not hinder new knowledge acquisition. This balanced trajectory confirms that UFSB successfully masters the new sketch-face task while simultaneously retaining high fidelity on the previous pedestrian task.

\textit{Performance Trends under Scheme B.}
Scheme B replaces CUFSF with the more challenging IIIT-D Viewed Sketch to evaluate robustness against natural style and pose variations. The corresponding trends are shown in Figs. \ref{fig:B_ms1k} and \ref{fig:B_iiit}.

In Fig. \ref{fig:B_ms1k}, the anti-forgetting capability of our method is further highlighted under this harder setting. Similar to Scheme A, baselines like LwF and DASK experience drastic collapses (LwF mAP drops $>65\%$, DASK R@1 $<1\%$). Conversely, UFSB maintains consistent performance close to its peak level even after completing the complex face sketch training. This stability underscores the robustness of our conformal prediction-based sample selection, which ensures that critical pedestrian features are preserved despite the high variability introduced by the IIIT-D dataset.

Fig. \ref{fig:B_iiit} reveals a particularly compelling finding on the IIIT-D dataset. While the Joint upper bound (trained on all data simultaneously) shows steady improvement, it eventually plateaus. Surprisingly, our sequential learning framework not only matches but \textit{surpasses} the Joint method in both mAP and R@1 by the end of training. This is a significant result: despite lacking access to global data distribution, UFSB leverages the synergistic effects of the SIM module (for cross-modal alignment) and MPM module (for knowledge preservation) to capture intrinsic, task-shared discriminative features more effectively than the non-sequential upper bound. This trend validates that our unified framework offers a superior solution for practical multi-task systems, achieving optimal generalization even in challenging, real-world scenarios.

\begin{figure}[t]
    \centering
    \subfloat[{\footnotesize UFSB - MaSk1K}]{\includegraphics[width=0.48\linewidth]{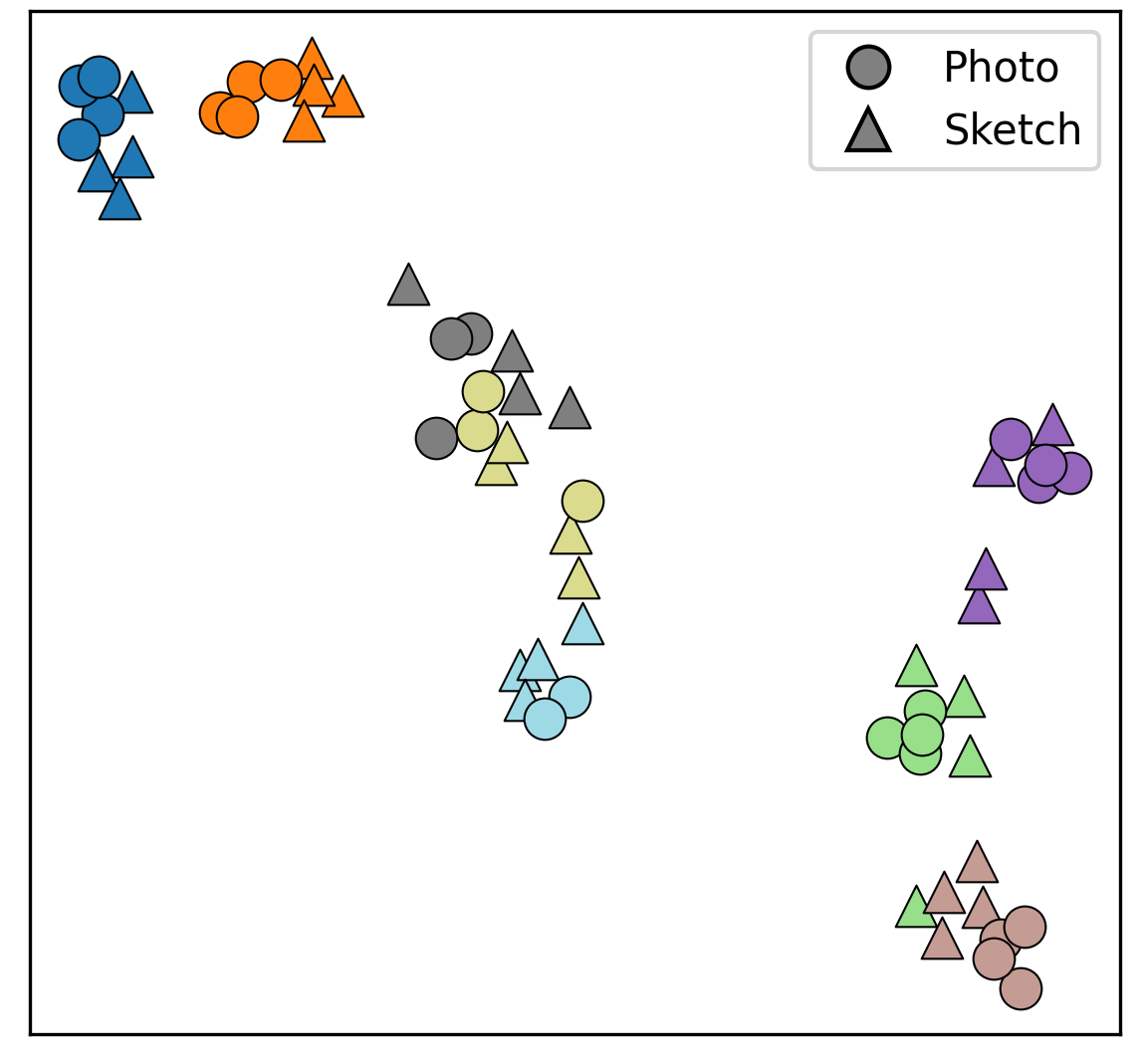}} \hfill
    \subfloat[{\footnotesize UFSB - CUFSF}]{\includegraphics[width=0.48\linewidth]{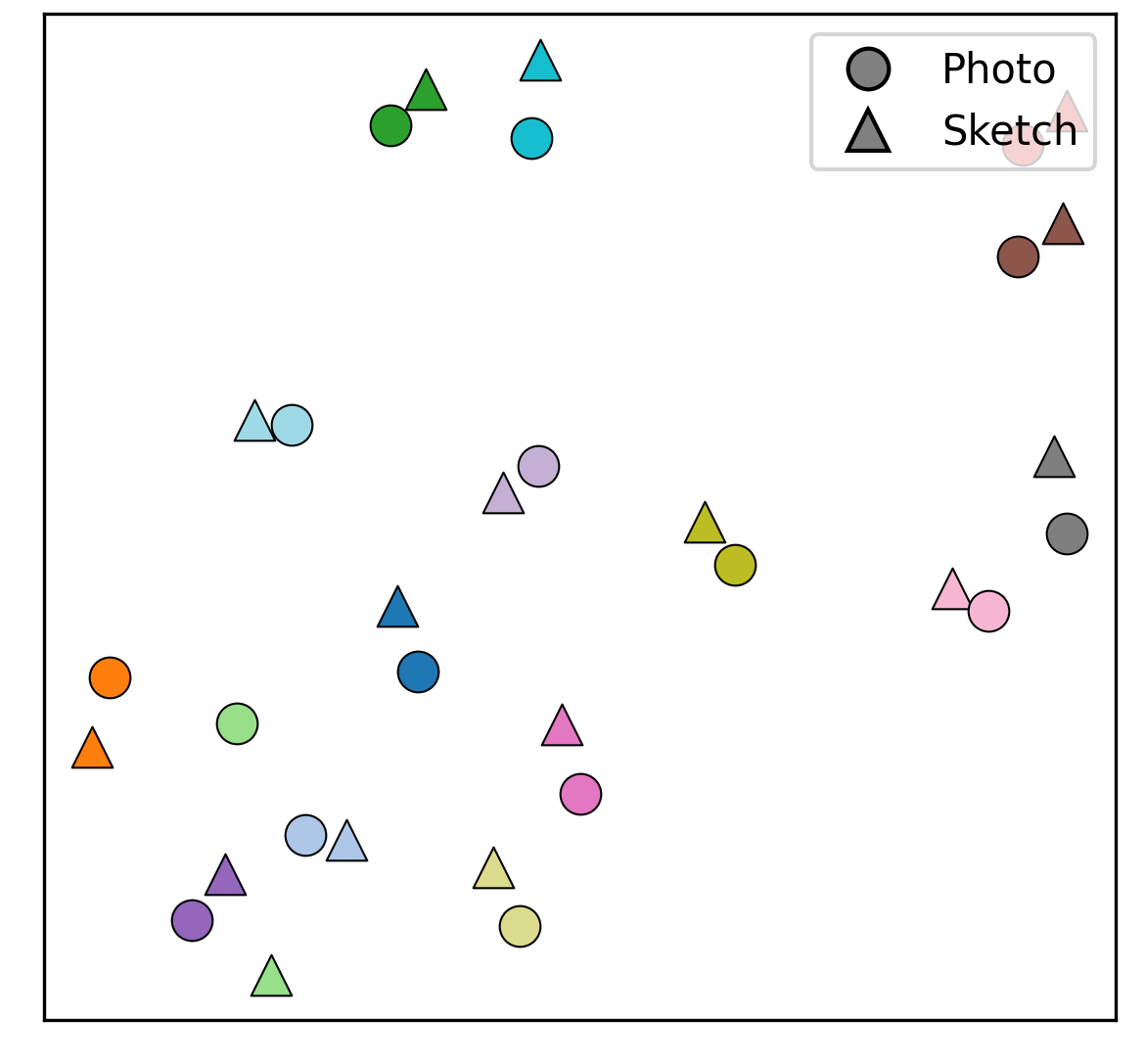}} \\
    \subfloat[{\footnotesize LwF - MaSk1K}]{\includegraphics[width=0.48\linewidth]{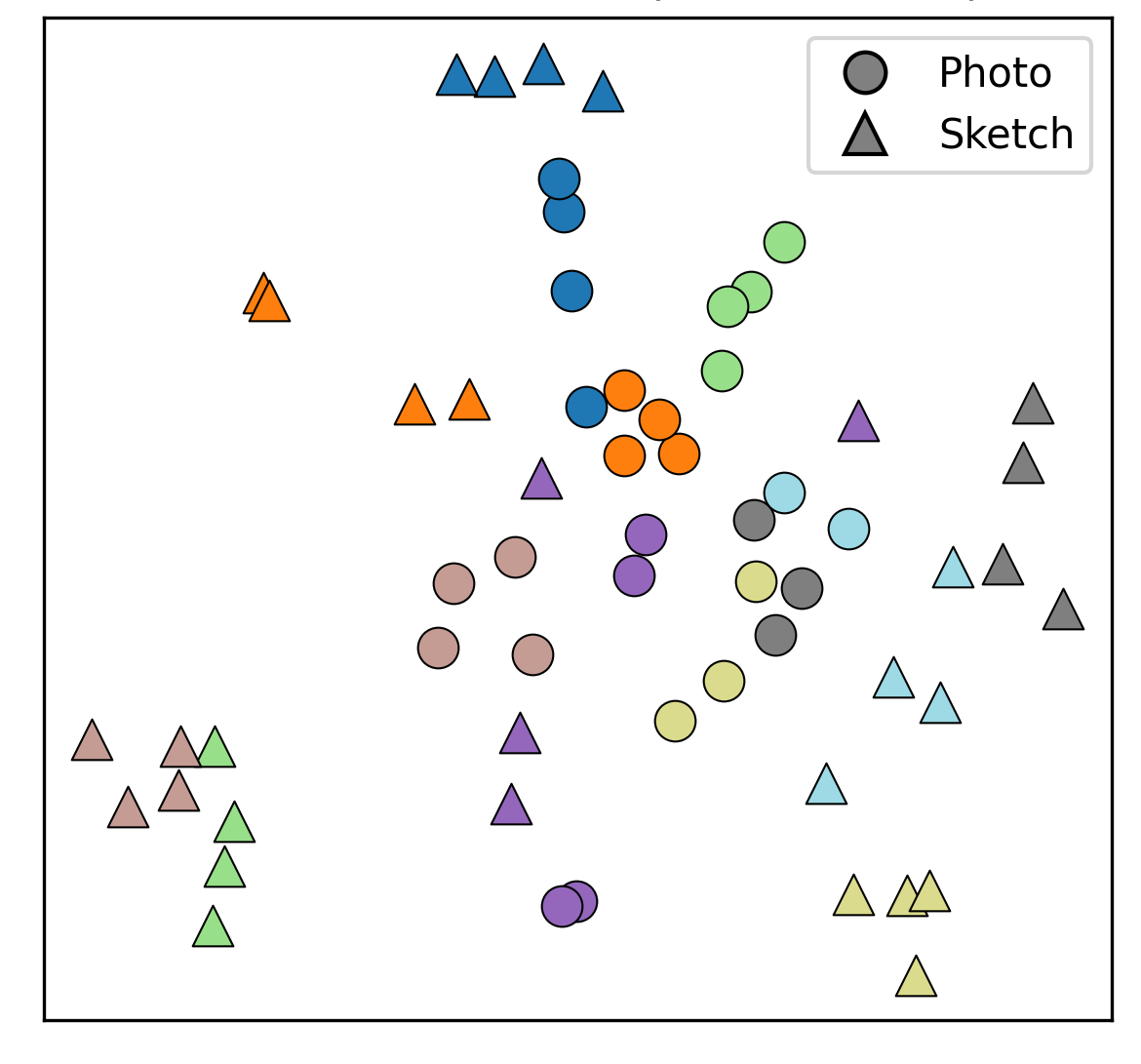}} \hfill
    \subfloat[{\footnotesize LwF - CUFSF}]{\includegraphics[width=0.48\linewidth]{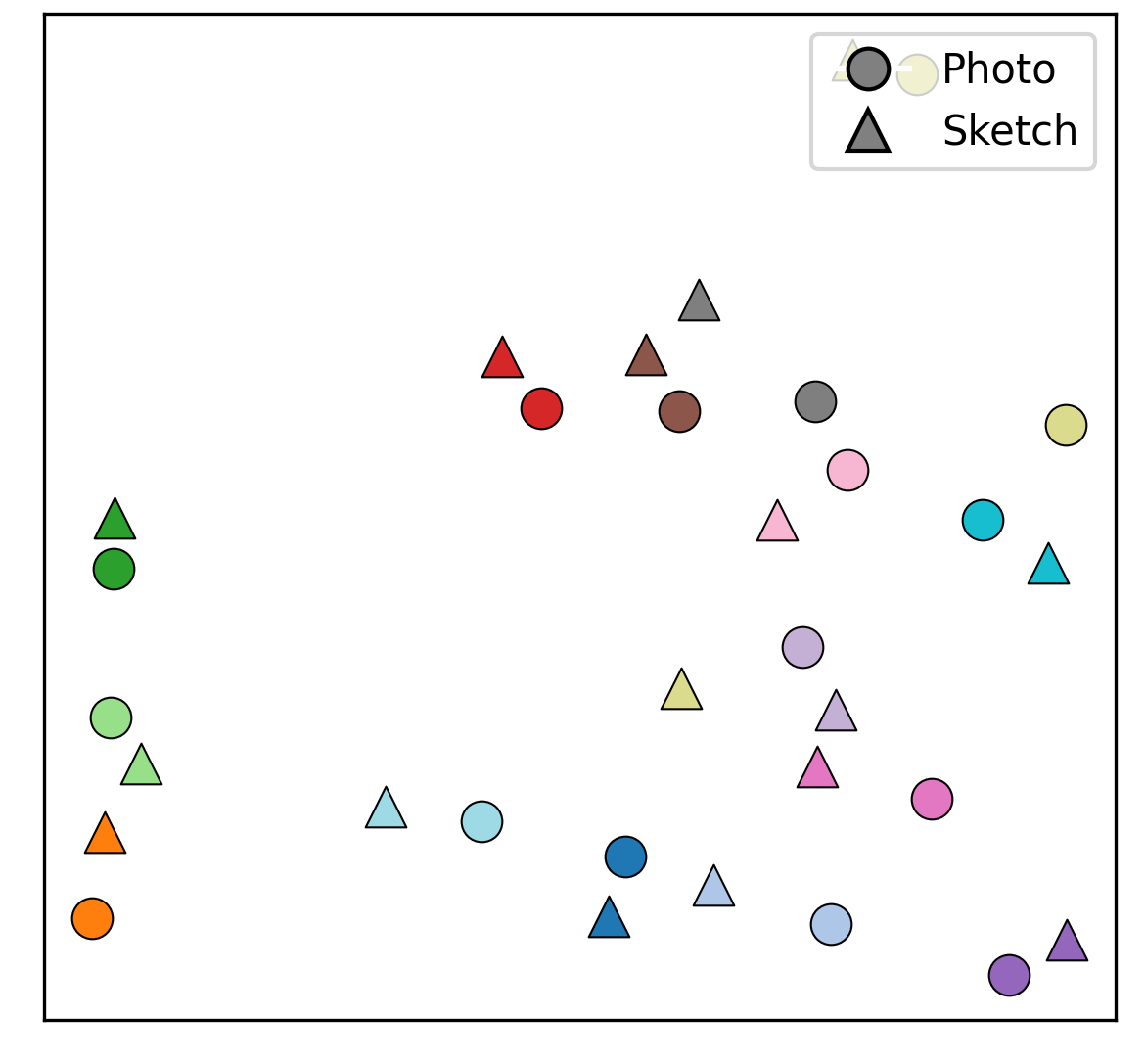}}
    \caption{t-SNE visualization of feature embeddings after sequential learning. Each column corresponds to a task (Person left, Face right); each row corresponds to a method (UFSB top, LwF bottom). Circles: Photo; Triangles: Sketch; Colors: identities. UFSB preserves compact, modality-aligned clusters for both tasks, while LwF suffers from severe forgetting and disentanglement.}
    \label{fig:tsne}
\end{figure}

To qualitatively assess the feature representation quality and anti-forgetting capability of our framework, we visualize the learned embeddings using t-SNE on both Task 1 and Task 2 test sets after Scheme A training. As shown in Fig.~\ref{fig:tsne}, we compare our UFSB with the representative continual learning baseline LwF.

In Fig.~\ref{fig:tsne}, the top row displays the feature distributions learned by our UFSB, while the bottom row shows those from LwF. For the Person task (left column), UFSB maintains well-separated clusters for each identity, with photo and sketch modalities tightly aligned within each cluster — indicating strong cross-modal alignment and preserved discriminative power for the old task. In contrast, LwF exhibits severe cluster overlap and modality disentanglement, confirming its susceptibility to catastrophic forgetting. Similarly, for the Face task (right column), UFSB achieves compact intra-class grouping and clear inter-class separation across both modalities, whereas LwF’s features are scattered and poorly structured. This visual evidence corroborates our quantitative results: our conformal prediction-based sample replay not only mitigates forgetting but also fosters a unified, task-shared embedding space where sketch-photo pairs from different domains coexist harmoniously without interference.

\subsection{Ablation Study}

\begin{table}[t]  
  \caption{Ablation Performance of w/o MPM and UFSB on Scheme A.}
  \label{tab:mpm_compare}
  \centering
  \resizebox{0.98\linewidth}{!}{
  \begin{tabular}{@{}lcccccc@{}}  
    \toprule
    Method & \multicolumn{2}{c}{MaSk1K} & \multicolumn{2}{c}{CUFSF} & \multicolumn{2}{c}{Avg} \\
    \cmidrule(lr){2-3} \cmidrule(lr){4-5} \cmidrule(lr){6-7}  
    & mAP & R@1 & mAP & R@1 & mAP & R@1 \\
    \midrule
    w/o MPM & 20.39 & 25.27 & 96.1 & 93.52 & 58.25 & 59.40 \\
    UFSB(Ours) & \textbf{27.19} & \textbf{31.69} & \textbf{96.59} & \textbf{94.24} & \textbf{61.89} & \textbf{62.97} \\
    \bottomrule
  \end{tabular}
  }
\end{table}

\begin{table}[t]
  \caption{Ablation performance of w/o MPM and UFSB on Scheme B.}
  \label{tab:mpm_iiitd_compare}
  \centering
  \resizebox{0.98\linewidth}{!}{
  \begin{tabular}{@{}lcccccc@{}}  
    \toprule
    Method & \multicolumn{2}{c}{MaSk1K} & \multicolumn{2}{c}{IIIT-D Sketch} & \multicolumn{2}{c}{Avg} \\
    \cmidrule(lr){2-3} \cmidrule(lr){4-5} \cmidrule(lr){6-7}  
    & mAP & R@1 & mAP & R@1 & mAP & R@1 \\
    \midrule
    w/o MPM & 24.26 & 29.49 & 85.89 & 80.71 & 55.08 & 55.10 \\
    UFSB(Ours) & \textbf{25.19} & \textbf{29.54} & \textbf{88.26} & \textbf{85.71} & \textbf{56.73} & \textbf{57.63} \\
    \bottomrule
  \end{tabular}
  }
\end{table}

To validate the MPM module, we conduct ablation studies on Scheme A and Scheme B, comparing our complete method (UFSB) with a variant excluding MPM (w/o MPM). The result is shown in the Table \ref{tab:mpm_compare} and Table \ref{tab:mpm_iiitd_compare}. Under Scheme A, our complete method yields substantial improvements on the old MaSk1K task: its mAP increases by 6.8\% and R@1 rises by 6.42\%. Meanwhile, it brings only minor gains to the new CUFSF task. This indicates MPM effectively mitigates catastrophic forgetting of the old pedestrian task without interfering with learning the new face task, and our complete method achieves better average metrics as a result. Scheme B is more challenging, using IIIT-D Viewed Sketch as the new task. Our method not only stabilizes performance on the old MaSk1K task but also boosts performance on the complex new IIIT-D task under this scheme: mAP rises by 2.37\% to 88.26\% and R@1 jumps by 5.00\% to 85.71\%. 

Collectively, these cross-scheme results verify MPM’s critical value: its replay mechanism not only alleviates catastrophic forgetting of old tasks but also enhances feature generalization for complex new tasks via retained discriminative knowledge, making it a key component for balancing sequential task performance in lifelong sketch-based biometric identification.

To validate synthetic data efficacy, we compare w/o Unreal and Ours under Scheme B (Table \ref{tab:unreal_compare}). Ours achieves 1.17\% higher mAP and 2.40\% higher R@1 on MaSk1K, and 15.71\% higher mAP and 15.00\% higher R@1 on IIIT-D Viewed Sketch. This boost stems from synthetic data supplementing scarce real sketches, covering edge cases and enhancing cross-modal feature generalization, effectively alleviating data scarcity and improving style robustness.

\begin{table}[t]
  \caption{Performance comparison of w/o Unreal and UFSB on Scheme B.}
  \label{tab:unreal_compare}
  \resizebox{0.98\linewidth}{!}{
  \centering
  \begin{tabular}{@{}lcccccc@{}}  
    \toprule
    Method & \multicolumn{2}{c}{MaSk1K} & \multicolumn{2}{c}{IIIT-D Sketch}  & \multicolumn{2}{c}{Avg} \\
    \cmidrule(lr){2-3} \cmidrule(lr){4-5} \cmidrule(lr){6-7} 
    & mAP & R@1 & mAP & R@1 & mAP & R@1\\
    \midrule
    w/o Unreal & 27.08 & 32.11 & 70.35 & 67.14 & 48.72 & 49.63\\
    UFSB(Ours) & \textbf{28.25} & \textbf{34.51} & \textbf{86.06} & \textbf{82.14} & \textbf{57.16} & \textbf{58.33}\\
    \bottomrule
  \end{tabular}
  }
\end{table}

\begin{table}[t]
  \caption{Comparison results on the \textbf{Reverse Scheme A} (CUFSF $\to$ MaSk1K). This setup reverses the training order of the original Scheme A, adopting CUFSF for the initial sketch-face recognition task and MaSk1K for the subsequent sketch ReID task.}
  \label{tab:dataset_compare_A_reverse}
  
  \centering
  \resizebox{0.98\linewidth}{!}{
  \begin{tabular}{@{}lcccccc@{}}  
    \toprule
    Method & \multicolumn{2}{c}{CUFSF} & \multicolumn{2}{c}{MaSk1K} & \multicolumn{2}{c}{Avg} \\
    \cmidrule(lr){2-3} \cmidrule(lr){4-5} \cmidrule(lr){6-7}  
    & mAP & R@1 & mAP & R@1 & mAP & R@1 \\
    \midrule
    \(\mathrm{Joint}^{+}\) & \textbf{98.62} & \textbf{97.55} & \underline{26.78} & \underline{30.59} & \textbf{62.7} & \textbf{64.07} \\
    \(\mathrm{LWF}^{+}\) \cite{45lwf} & 89.01 & 84.01 & 5.38 & 6.5 & 47.2 & 45.26 \\
    DKP \cite{22dkp} & \underline{97.3} & \underline{95.8} & 3.7 & 4.4 & 50.5 & 50.1\\
    DASK \cite{23dask} & 36.9 & 28.2 & 2.7 & 2.7 & 19.8 & 15.5 \\
    \(\mathrm{UFSB(Ours)}^{+}\) & 97.02 & 95.24 & \textbf{26.83} & \textbf{32.11} & \underline{61.95} & \underline{63.68} \\
    \bottomrule
  \end{tabular}
}
\end{table}

\textit{Robustness to Task Order Variation}
To further evaluate the robustness of our framework against task sequence variations, we conducted an additional experiment denoted as \textbf{Reverse Scheme A}, where the training order is inverted compared to the standard Scheme A. Specifically, the model is first trained on the sketch-face recognition task (CUFSF) and subsequently on the sketch-pedestrian ReID task (MaSk1K). The quantitative results are presented in Table \ref{tab:dataset_compare_A_reverse}.

As observed in Table \ref{tab:dataset_compare_A_reverse}, existing lifelong ReID methods exhibit severe sensitivity to task ordering. When the complex face recognition task is learned first, followed by the pedestrian ReID task, methods like LwF and DKP suffer from catastrophic forgetting on the initial task (CUFSF). For instance, LwF's performance on CUFSF drops significantly (mAP: $89.01\%$, R@1: $84.01\%$) compared to the Joint upper bound, while DKP almost completely loses its discriminative capability on the first task (mAP: $3.7\%$). This indicates that their knowledge preservation mechanisms fail to generalize across different task domains and orders.

In stark contrast, our UFSB framework demonstrates remarkable stability regardless of the task sequence. Even when facing the challenging transition from face to pedestrian sketches, UFSB retains high performance on the old task (CUFSF: $97.02\%$ mAP, $95.24\%$ R@1), closely approaching the Joint upper bound and significantly outperforming all other continual learning baselines. Simultaneously, it achieves state-of-the-art performance on the new task (MaSk1K: $26.83\%$ mAP, $32.11\%$ R@1), \textit{surpassing the Joint method} in pedestrian identification metrics. The average performance of UFSB ($61.95\%$ mAP) remains highly competitive with the Joint upper bound ($62.70\%$ mAP), validating that the synergistic effect of our SIM and MPM modules is not dependent on a specific task order. This result confirms that UFSB effectively captures task-agnostic shared representations, ensuring reliable biometric identification across diverse sequential learning scenarios.

Performance tendencies in Figs.\ref{fig:cufsf} and Figs.\ref{fig:iiit} further validate our framework’s stability, anti-forgetting capability, and adaptive learning ability. Unlike lifelong ReID methods that suffer from catastrophic forgetting, our method maintains consistent or even improved performance across both tasks. After completing training on CUFSF or IIIT-D Viewed Sketch, its performance on MaSk1K remains stable, and its performance on the new task continues to improve. This stability and adaptability originate from the MPM module’s trusted sample replay, which not only preserves pedestrian-specific discriminative knowledge without interfering with new task learning but also enables synergistic learning between pedestrian and face tasks. More detailed comparison plots of performance trends between our method and other LReID methods under the two schemes are presented in \emph{supplementary material}.

\textit{Key takeaways from comparisons.}
\emph{1.~Cross-Modal Superiority:} existing LReID methods fail to handle sketch-photo matching due to their lack of cross-modal optimization. Our framework addresses this via the SIM module’s JMMD constraint, achieving competitive cross-modal performance.
\emph{2.~Cross-Task Generalization:} unlike single-task cross-modal ReID methods, our unified framework simultaneously supports sketch-pedestrian and sketch-face ReID, with average mAP/R@1 surpassing LReID methods by 11.23\%–35.18\%.
\emph{3.~Synthetic Data Value:} fusing synthetic data consistently improves performance across tasks, validating its effectiveness in reducing data collection costs and mitigating privacy risks.
\emph{4.~Robustness to Challenging Scenarios:} on the difficult IIIT-D Viewed Sketch dataset, our framework outperforms even the upper-bound JointTrain method, demonstrating strong generalization to real-world sketch variations.
\emph{5.~Order Agnostic Robustness:} our framework maintains stability regardless of training order. By capturing task-agnostic shared representations, it retains high fidelity on old tasks while achieving state-of-the-art performance on new ones, confirming reliable identification across diverse sequential scenarios.

\section{Conclusion}
In this paper, we focus on the unified cross-modal sketch biometric identification task, addressing core challenges like scarce real sketch data, high annotation costs, and catastrophic forgetting in cross-task lifelong learning. To solve these, we propose a unified framework integrating synthetic data generation, cross-modal alignment, and multi-task preservation: we construct the SketchUnified-BioID benchmark, design two core modules, the MPM module for low-uncertainty sample replay via conformal prediction to mitigate forgetting, and train tasks sequentially to enable incremental multi-task knowledge accumulation. Extensive experiments show our method outperforms SOTA. With ablations confirming MPM’s anti-forgetting effect and synthetic data’s generalization boost. In future work, we will extend the framework to other sketch-based biometrics such as palmprint and fingerprint sketches. We believe our work provides a new paradigm for unified cross-modal biometric identification and inspires further research in lifelong sketch re-identification and other biometrics tasks.




\bibliographystyle{IEEEtran}
\bibliography{main}

\vfill

\end{document}